\documentclass[runningheads]{llncs}
\usepackage{graphicx,wrapfig}
\usepackage{subcaption}
\usepackage{wrapfig}
\usepackage{geometry}
\usepackage{amsmath}
\usepackage{hyperref}
\usepackage{adjustbox}

\begin{document}
\title{Habitual and Reflective Control in Hierarchical Predictive Coding}
%
%
\author{Paul F Kinghorn\inst{1}, Beren Millidge\inst{2}\and
Christopher L Buckley\inst{3}}
\authorrunning{PF Kinghorn, B Millidge and CL Buckley}
\institute{School of Engineering and Informatics,
University of Sussex \email{p.kinghorn@sussex.ac.uk} 
\and MRC Brain Networks Dynamics Unit,
University of Oxford \email{beren@millidge.name}
\and School of Engineering and Informatics,
University of Sussex \email{c.l.buckley@sussex.ac.uk}
}
\maketitle              
\begin{abstract}
In cognitive science, behaviour is often separated into two types. Reflexive control is habitual and immediate, whereas reflective is deliberative and time consuming. We examine the argument that Hierarchical Predictive Coding (HPC) can explain both types of behaviour as a continuum operating across a multi-layered network, removing the need for separate circuits in the brain. On this view, “fast” actions may be triggered using only the lower layers of the HPC schema, whereas more deliberative actions need higher layers. We demonstrate that HPC can distribute learning throughout its hierarchy, with higher layers called into use only as required.

\end{abstract}
\section{Introduction}
In the field of cognitive science, behaviour is widely considered to be separated into two classes. Habitual (reflexive) behaviour responds rapidly and instinctively to stimuli, while reflective behaviour involves slower top-down processing and a period of deliberation. 
These different classes of behaviour have been labelled System 1 and System 2 by Stanovich and West in 2000~\cite{stanovich2000individual} and the topic was  popularised in Daniel Kahneman’s book "Thinking Fast and Slow" in 2011~\cite{KahnemanDaniel2012Tfas}. However, it has remained unclear how the two processes are implemented in the brain.

This paper investigates how a single system operates when generating behaviours which require different amounts of deliberation.  The system does not carry out planning or evaluation of prospective outcomes, but simply compares the triggering of actions which require more or less time to select the correct action for a presented situation. This distinction has some parallels with, but is separate from, the split between goal based planning (often called model-based) and habitual (often called model-free) control~\cite{DolanRayJ2013GaHi,SuttonRichardS2018Rl:a}. Although there is some evidence that separate brain systems underpin goal based planning and habits~\cite{glascher2010states,DolanRayJ2013GaHi,WunderlichKlaus2012Mvbp}, there are also indications from fMRI and lesion studies that these processes can co-exist within the same regions of the brain~\cite{daw2011model,DolanRayJ2013GaHi,WunderlichKlaus2012Mvbp}, challenging the notion of separate systems. Since it is possible that even these extremes of behaviour are computed together in the brain, a useful contribution to the topic would be to analyse how reflexive and reflective behaviour can arise from a unified system.

Independent from this debate, there has been much progress in the last decade on the idea that perception and action are both facets of the principle of free energy minimization in the brain~\cite{friston2006free,FristonKarl2010Tfpa}. According to this approach, the brain maintains a generative model of its environment and uses variational inference to approximate Bayesian inference~\cite{SethAnilK2015TcBb,hohwy2013predictive}. 
One way of implementing this 
(under Gaussian assumptions) is to use a hierarchical predictive coding architecture~\cite{rao1999predictive,friston2003learning,friston2005theory}, which has successive layers of descending predictions and ascending prediction errors~\cite{BuckleyChristopherL2017Tfep,bogacz2017tutorial,whittington2017approximation,millidge2019combining}. This theory is also often referred to as Predictive Processing (PP)~\cite{ClarkAndy2013WnPb,PezzuloGiovanni2017MAtA}. 
In this paper, we investigate how PP approaches can explain both reflexive and reflective behaviour simultaneously using a single hierarchical predictive coding network architecture and inference procedure.\\
\indent
In it basic form, PP uses a generative model to try and correctly infer hidden causes for incoming observations. In a hierarchical predictive coding network, all layers of the hierarchy are updated to minimize prediction errors until a fixed point is reached, with the resultant top layer being the best explanation of the hidden causes of the observations at the bottom layer~\cite{ClarkAndy2013WnPb,BuckleyChristopherL2017Tfep,bogacz2017tutorial}.

\indent PP can also be used to explain action, with the network modelling how actions and sensations interact~\cite{BaltieriManuel2019Gmap,burr2016embodied,BruinebergJ2018Tabi,clark2015predicting,RamsteadMaxwellJD2020Atot,PezzuloGiovanni2015AIhr,PezzuloGiovanni2017MAtA}. 
 Actions are triggered by descending predictions which cause low level prediction errors. These errors are rectified through reflex arcs~\cite{friston2010action,adams2013predictions,hipolito2021embodied}. In theory, this means that motor behaviour need not wait for full end-to-end inference to be completed but, rather, action takes place once a threshold has been crossed on the reflex muscle. 
 
\indent
This paper investigates the extent to which action selection in a predictive coding network (PCN) relies on all the layers of the PCN. To do this, we train a network to associate actions and observations 
with each other. We then investigate whether inference across the full network is required in order to trigger the correct action for a given observation. We show that a decision making task with a higher degree of complexity will use more of the layers and may be strongly dependent on the top layer being correctly inferred. Conversely, a decision which is a simple function of sensory observations can operate without involvement of higher layers, despite the fact that learning included those higher layers. This demonstrates that learning allows a hierarchy of action/sensation linkages to be built up in the network, with agents able to use information from lower layers to infer the correct actions without necessarily needing to engage the whole network. These findings suggest that a single PCN architecture could explain both reflexive and reflective behaviour.

In the general case of state space models, the fixed point of a PCN is often in a moving frame of reference. However, the implementation described in this paper ignores state transitions or dynamics and restricts itself to static images. It should therefore not be confused with the notion of predictive coding sometimes seen in the engineering or active inference literature which rests on a state space model for generating timeseries. Rather, our formulation follows the approach of Rao and Ballard's seminal paper~\cite{rao1999predictive} and ignores any temporal prediction components, whilst retaining what Friston describes as "the essence of predictive coding, namely any scheme that finds the mode of the recognition density by dynamically minimising prediction error in an input-specific fashion"~\cite{friston2003learning}.

The remainder of this paper is set out as follows. Section 2 outlines the HPC model which is used to implement variational inference.  Section 3 describes the experiments which we use to analyse inference of labels and actions in PCNs. Section 4 presents the experimental results, demonstrating that learning to act need not rely on high level hidden states. Moreover, we show that the number of higher layers which can be ignored in decision making relates to the complexity of information needed to make that decision.

\section{Hierarchical Predictive Coding (HPC)} 
\label{HPC_section}
This section presents a quick overview of how HPC can be used to approximate variational inference. For a more guided derivation, see~\cite{BuckleyChristopherL2017Tfep,bogacz2017tutorial,millidge2019combining,friston2006free}.

At the core of the free-energy principle is the concept that, in order to survive, an agent must strive to make its model of the world a good fit for incoming observations, \(o\). If the model of observations \(p(o)\) can be explained by hidden states of the world \(s\) then, in theory, a posterior estimate for \(s\) could be obtained using Bayes rule over a set of observations:

\begin{equation}
\label{bayes}
p(s\mid o) 
=\frac{p(o \mid s) \ p(s)}{p(o)}
= \frac{p(o \mid s) \ p(s)}{\int {p(o \mid s) \ p(s)}\ ds}
\end{equation}
 but the denominator is likely to be intractable. Therefore \(s\) is approximated using variational inference.  An auxiliary model (the variational distribution) is created, \(q(s; \psi)\), and the divergence between \(q\) and the true posterior \(p(s \mid o)\) minimized. The KL divergence is used to measure this: 
\begin{equation}
\label{KL}
KL[q(s; \psi) \parallel p(s\mid o) ]
=
\int{q(s; \psi) log\frac{q(s; \psi)}{p(s\mid o)}}ds
=
\mathcal{F} \\ + \\ log \ p(o)
\end{equation}
where the variational free energy \(\mathcal{F}\) is defined as: 
\begin{equation}
\mathcal{F} =\int{q(s; \psi) log\frac{q(s; \psi)}{p(s , o)}}ds
\end{equation}

The value \(-log \ p(o)\), is an information theoretic measure of the unexpectedness of an observation, variously called surprise, suprisal or negative of log model evidence. By adjusting \(s\) to minimize surprisal, the model becomes a better fit of the environment. Noting that KL is always positive, it can be seen from equation~\eqref{KL} that \(\mathcal{F}\) is an upper bound on surprisal. Therefore, to make the model a good fit for the data, it suffices to minimize \(\mathcal{F}\).   

The next step is to consider how this would be implemented in the brain via HPC. 
In HPC, the generative model \(p(s , o)\) is implemented in Markovian hierarchical layers, where the priors are simply the values of the layer above, mapped through a weight matrix and a nonlinear function. The prior at the top layer may either be a flat prior or set externally. With \(N\) layers, the top layer is labelled as layer \(1\), and the observation at the bottom as layer \(N\). Thus:
\begin{equation}
\label{Markov}
p(s , o) = p(s_{N} \mid s_{N-1})\dots p(s_{2} \mid s_{1}) \  p(s_{1}) \ \textrm{where} \ s_N = o
\end{equation}

The generative model is assumed to be Gaussian at each layer, 
\begin{equation}
p(s_{n+1} \mid s_{n}) = N (s_{n+1}  ;  f(\Theta_n \  s_{n}) , \Sigma_{n+1})
\end{equation}
where \(s_{n}\) is a vector representing node values on layer n, \(\Theta_{n}\) is a matrix giving the connection weights between layer \(n\) and layer \(n+1\), \(f\) is a non-linear function and \(z_{n}\) is Gaussian noise at each layer. Note that the network also has a bias at each layer which is updated in a similar manner to the weights. This has not been included here for brevity. [Here we have shown the form where the argument of \(f\) is a weighted linear mixture of hidden states, in order to make clear how we have implemented the hierarchy. But this could equally be generalised to any non linear function \(f\).]

Making the assumption that \(q\) is a multivariate Gaussian distribution \(q(s) \sim N(s ; \mu, \Sigma )\), 
and further assuming that the distribution of \(q\) is tightly packed around \(\mu\) (to enable use of the Laplace assumption), 
\(\mathcal{F}\) reduces to:  
\begin{equation}
\label{F_approx}
\mathcal{F} \approx - log \ p(\mu , o)
\end{equation}
where \(o\) is a vector representing observations and \(\mu\) is the mean of the brain's probability distribution for \(s\). It is important to note that in this paper the observations are not confined to incoming senses but also include actions, in the form of proprioceptive feedback. Exteroceptive observations cause updates to model beliefs which, in turn, result in updated beliefs on proprioceptive observations. These drive motoneurons to eliminate any prediction error through reflex arcs~\cite{friston2010action,friston2011optimal,SethAnilK2015TcBb}. Action can therefore be thought of as just a particular type of observation.

Using the distribution for a multivariate Gaussian, the estimate of \(\mathcal{F}\) can be transformed into:

\begin{equation}
\label{F2}
\begin{split}
\mathcal{F} \approx -log \ p(\mu , o) 
= \sum_{n} log \ p(\mu_{n+1} \mid \mu_{n}) 
= \sum_{n} -\frac{1}{2}  \epsilon_{n+1}^T \Sigma_{n+1} ^{-1} \epsilon_{n+1} - \frac{1}{2} log (2 \pi \ | \Sigma_{n+1} |)  
\end{split}
\end{equation}
where \(\epsilon_{n+1} := \mu_{n+1}-f(\Theta_n \ \mu_{n} )\) is the difference between value of layer \(n+1\) and the value predicted by layer \(n\).
\(\mathcal{F}\) is then minimized following the Expectation-Minimization approach~\cite{dempster1977maximum,mackay2003information}, by using gradient descent to alternately update node values (\(\mu\)) on a fast timescale and weight values (\(\Theta\)) on a slower timescale.

The gradient for node updates in a hidden layer uses the values of \(\epsilon_{n}\) and \(\epsilon_{n+1}\), and is given by the partial derivative:

\begin{equation}
\label{updates2}
\frac{\partial\mathcal{F}}{\partial\mu_n}=
\epsilon_{n+1} \  
\Sigma_{n+1}^{-1} \  \Theta_n^T \ 
f'(\Theta_n \ \mu_n)  - 
\epsilon_{n} \ 
\Sigma_{n}^{-1} 
\end{equation}
but if the node values of the top layer are being updated then this is truncated to only use the difference compared to the layer below: 
\begin{equation}
\label{updates3}
\frac{\partial\mathcal{F}}{\partial\mu_1}=
\epsilon_{2} \
\Sigma_{2}^{-1} \  \Theta_1^T \
f'(\Theta_1 \ \mu_1) 
\end{equation}

As pointed out earlier, downward predictions not only predict exteroceptive (sensory) signals, but also create a proprioceptive prediction error in the motor system (which is cancelled by movement via a reflex arc).  In this paper we simply intend to monitor the signals being sent to the motor system and do not wish to include the error cancellation signal being fed back from the reflex arc. For this reason, the update of the "observation node" in the motor system is shown as only using the difference to the layer above:

\begin{equation}
\label{updates4}
\frac{\partial\mathcal{F}}{\partial\mu_N}=
 - 
\epsilon_{N} \
\Sigma_{N}^{-1} 
\end{equation}

After the node values have been changed, \(\mathcal{F}\) is then further minimized by updating the weights using:

\begin{equation}
\label{updates1}
\frac{\partial\mathcal{F}}{\partial\Theta_n} =
{\epsilon_{n+1} \ \Sigma_{n+1}^{-1}} \
\mu_n^T \ 
f'(\Theta_n \ \mu_n) 
\end{equation}

Since the impact of variance is not the primary focus here, our simulations assume that all \(\Sigma\) have fixed values of the identity matrix and therefore the gradient update for \(\Sigma\) has not been included. 

Fig.~\ref{digit1details:test_mode} summarises the flow of information in the network during gradient descent update of node values. 

\section{Methods}
Three sets of experiments were designed, to investigate how the process of inference is distributed through hierarchical layers. The first two experiments were each run on three different tasks. The third experiment was run on a single task. The experiments are described below.

\begin{figure}
    \centering
    \begin{tabular}{cc}
    \adjustbox{valign=b}{\subfloat[Network in test mode\label{digit1details:test_mode}]{%
          \includegraphics[width=.45\linewidth,height=6cm]{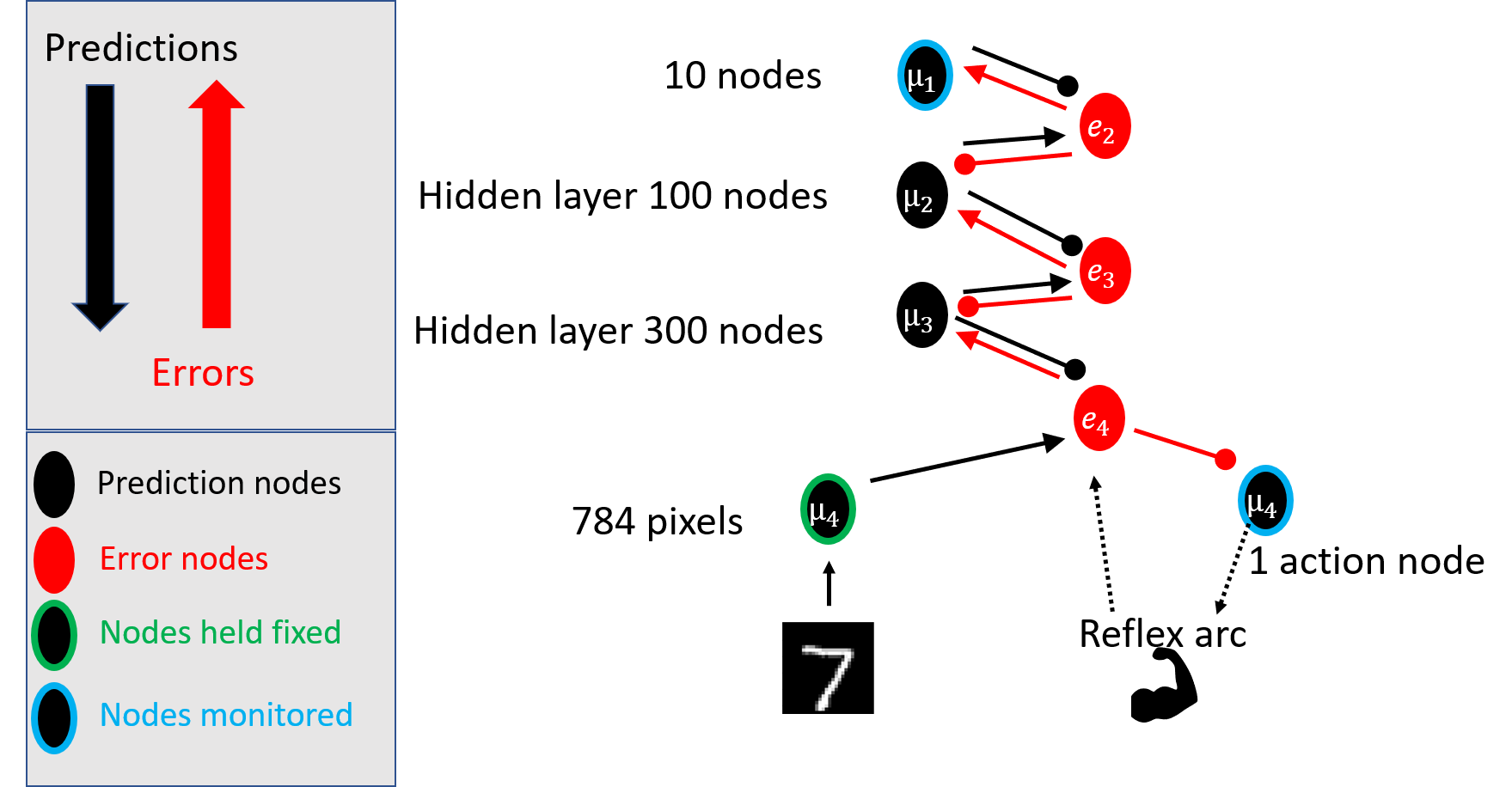}}}
    &      
    \adjustbox{valign=b}{\begin{tabular}{@{}c@{}}
    \subfloat[Distribution of inferred actions\label{digit1details:digit1_distribution}]{%
          \includegraphics[width=.45\linewidth,height=3cm]{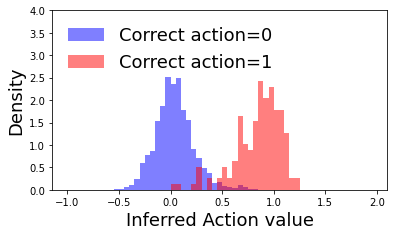}} \\
    \subfloat[Development of label and action accuracy\label{digit1details:digit1_development}]{%
          \includegraphics[width=.45\linewidth,height=3cm]{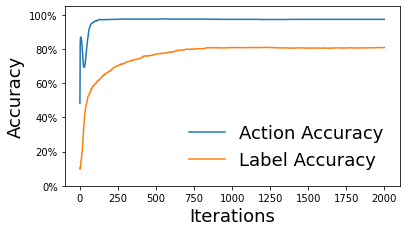}}
    \end{tabular}}
    \end{tabular}
    \caption{Results for MNIST-digit1:  correct action=1 if label=1, or  0 otherwise. (a) Test configuration. The observation nodes representing the MNIST picture are held fixed and both the label and the action nodes are updated using variational inference. (b) Distribution of inferred actions at end of inference period.  (c) Simultaneous development of accuracy for label and action as inference progresses through iterations. Using argmax on label nodes, label accuracy = 80.8\%. Using a heaviside function centred on 0.5, action accuracy = 97.4\%.  This indicates that the correct label is not required in order to select the correct action.}
    \label{digit1details}
\vspace{-0.65cm}
  \end{figure}

In the first set of experiments, we trained three PCNs to carry out separate inference tasks, based on selecting the correct action for a given MNIST image~\cite{lecun-mnisthandwrittendigit-2010}.  In all three networks, the observation layer at the bottom of the network contains 785 nodes, made up of 784 sensory nodes (representing the pixels of an MNIST image) and a single binary action node. 
The top layer uses a one-hot representation of each of the possible MNIST labels. There are two hidden layers of size 100 and 300. Thus, if there are 10 possible labels, there is a four-layer network of size [10,100,300,785], whose generative model produces an MNIST image and an action value from a given MNIST label. The role of each of the networks is, on presentation of an MNIST image, to infer the correct MNIST label at the top and the correct action associated with that image (Fig.~\ref{digit1details:test_mode}).  

We investigated  the relationship between the accuracies of action inference and label inference. Specifically, we asked: to what extent can the action be correctly triggered without correct label inference?

In the first task, MNIST-digit1, we trained the action node to output value 1 if the presented MNIST image has label 1, and value 0 for all other digits, i.e. the job of the action node is to fire when an image of the digit 1 is presented. The network is trained in a supervised manner to learn the generative model, 
by fixing the top and bottom layers with the training labels and observations respectively, and then, minimizing \(\mathcal{F}\) in an expectation–maximization (EM) fashion~\cite{dempster1977maximum,mackay2003information}, as described in Section~\ref{HPC_section}. Once trained, the network is then tested for its ability to infer the correct label and action for a given image. This is done by presenting an MNIST image to the 784 sensory states and allowing both the labels at the top and the action at the bottom to update via the variational inference process, according to equations \eqref{updates2} -  \eqref{updates4}. Updates to the network are applied over a large number of iterations and, at any stage of this process, the current inferred label can be read out as the argmax of the top-layer nodes while the selected action is read out according to a heaviside function applied to the action node value, centred on 0.5.

In the second task, MNIST-groups, we trained the action node to fire if the MNIST label is less than 5, and not fire otherwise. This network is trained and tested using the same process as above.\\
\indent In the third task, MNIST-barred, half of the MNIST images had a white horizontal bar applied across the middle of the image. A new set of labels was created so that there were now 20 possible labels - 0 to 9 representing the digits without bars, and 10 to 19 representing the digits with bars. Action value 1 was associated with labels 10 to 19, and action value 0 with labels 0 to 9. The network for this task has size [20,100,300,785]. It is trained and tested as for the first two tasks. Appendix~\ref{appendix:network_params} gives full details of the hyperparameters used in the three PCNs.\\
\indent The second set of experiments used the same three tasks but, instead of fixing MNIST labels to the top of the network in training, the top layer was populated with random noise. The purpose of these experiments was to determine whether the provision of label information in training had any impact on the network's ability to infer the correct action. We then ablated layers from the PCNs in order to investigate the contribution which each layer makes towards inferring the correct action.

The third experiment trained a network where both the MNIST image and the MNIST one hot-labels were placed at the bottom. Above this were 6 layers, all initialized with noisy values. The top layer was allowed to vary freely (see Fig.~\ref{onehot network}). This was used to investigate how label inference performed in this scenario (rather than the traditional case of label at the top and image at the bottom), and how performance reacted to ablation of layers in test mode.

\section{Results}
\vspace{-0.2cm}
We first investigated the relationship between accuracy of action and label inference for MNIST-digit1 (where the action node should fire if the MNIST label is 1). When run on a test set of images, the network generates values on the action node which correctly split  into two groups centred near to 0 and 1, with a small overlap (Fig.~\ref{digit1details:digit1_distribution}). As a result, the network is able to correctly infer the action for a presented image in over 97\% of cases. On the other hand, the label is only correctly inferred in 81\% of cases, demonstrating that action selection does not depend entirely on correct label inference.
Fig.~\ref{digit1details:digit1_development} presents the development of label and action accuracies as iterations progress, confirming that a) action accuracy is always better than label accuracy, b) further iterations will not change this and c) action inference reaches asymptotic performance quicker than label inference.

\begin{wrapfigure}{r}{0.5\linewidth}
\vspace{0cm}
 \includegraphics[width=\linewidth]{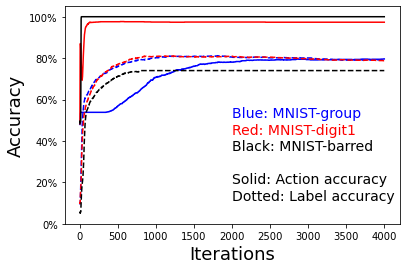}
\caption{Development of inference accuracy for label (at top of network) and action (at bottom of network), for 3 different tasks. Blue lines: MNIST-group. Action accuracy is constrained by label accuracy at around 80\%. Red lines: MNIST-digit1. Action selection is above 97\% accuracy, with label accuracy around 80\%. Black lines: MNIST-barred. Accuracy of action selection quickly reaches 100\% despite label accuracy being low. On a poorly trained network the results are even more striking, with action accuracy still perfect despite label accuracy of only 20\% (results not shown).}
\label{development}
\vspace{-0.5cm}
\end{wrapfigure}

Fig.~\ref{development} compares label and action accuracy for all three tasks. In the MNIST-group task, action accuracy appears to be constrained by label accuracy. In the MNIST-barred task, the correct action is always inferred, even though the network has relatively poor label accuracy. It would therefore seem that the MNIST-group task is reliant on upper layer values in order to select the correct action, whereas the simpler tasks can reach, or approach, optimal action performance regardless of the upper layer values.

However, it is not clear from these results whether the MNIST-group task is relying on the fact that the higher layers contain information about the image labels (recall that this is how the network was trained) or whether it is simply that the existence of the higher layers is providing more compute power. To investigate this, the second set of experiments were run, where the three networks are trained with random noise at the top layer instead of the image label. In testing, label accuracy was now no better than random (as one would expect), but action accuracy was indistinguishable from the original results of Fig.~\ref{development}.  This demonstrates that it is the existence of the layers, rather than provision of label information in training which is driving action inference.

\begin{figure}
 \vspace{-0.4cm}
 \captionsetup{justification=raggedright}
\centering
\begin{subfigure}{.33\textwidth}
  \centering
  \includegraphics[width=.9\linewidth]{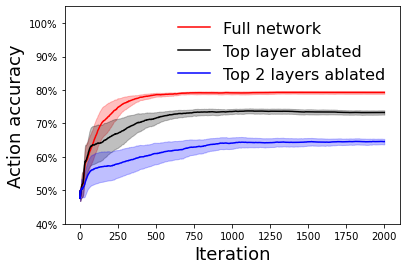}
  \caption{MNIST-group }
  \label{fig8:group}
\end{subfigure}%
\begin{subfigure}{.33\textwidth}
  \centering
  \includegraphics[width=.9\linewidth]{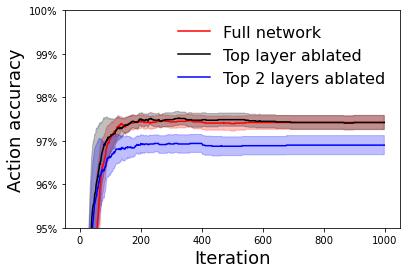}
  \caption{MNIST-digit1}
  \label{fig8:digit1}
\end{subfigure}
\begin{subfigure}{.33\textwidth}
  \centering
  \includegraphics[width=.9\linewidth]{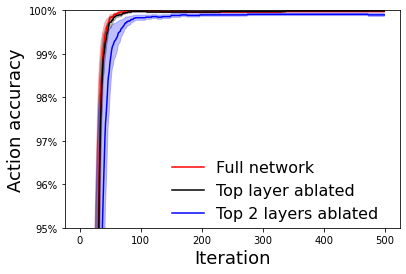}
  \caption{MNIST-barred}
  \label{fig8:barred}
\end{subfigure}
\caption{Effect of ablating layers on action accuracy.  The three tasks cope differently with ablation of layers, as shown in (a), (b) and (c). Note that different y-scales are used on the figures for clarity. Each network was trained using 6 different seeds, and error bars show standard error. Results suggest that, if the lower layers are sufficient for action selection then the higher layers can be ignored.}
\label{fig8}
\vspace{-0.4cm}
\end{figure}

To confirm that the three tasks make different use of the higher layers, action accuracy was measured when the top two layers were ablated in test mode (they were still present in training). Performance on the MNIST-group task (Fig.~\ref{fig8:group}) deteriorates significantly as the layers are ablated. Conversely, ablation of the top layer has no impact on the action accuracy of either the MNIST-barred (Fig.~\ref{fig8:barred}) or MNIST-digit1 tasks (Fig.~\ref{fig8:digit1}). Both suffer slightly if the top 2 layers are ablated, although in the case of MNIST-barred the accuracy only moves from 100\% to 99.9\%. 
It can be concluded from these ablation experiments that reliance on higher layers varies with the nature of the task. Tasks which are more challenging may rely on the higher layers, while simple tasks may not suffer at all if the layers are ablated - presumably because all the information required for action selection is entirely available in the lower layers.

In the third experiment, we constructed a network with both MNIST image and MNIST one
hot-labels at the bottom, representing 10 different binary actions to select from (see Fig.~\ref{onehot network}). Above this were 6 layers, all initialized with noisy values (details in Appendix~\ref{appendix:network_params}). Training was carried out as before, presenting a set of images and labels at the bottom of the network and leaving the network to learn weights throughout the hierarchy. The effect of layer ablation on the ability of the network to select the correct action (which in this experiment is the one-hot label) was then tested. When using all the layers, this network produces comparable results to the more standard PCN setup with label at the top and image at the bottom.\footnote{At approximately 78\%, the accuracy we achieved is significantly lower than standard non-PCN deep learning methods. This is partly because the model has not been fine-tuned (e.g. hyper-parameters, using convolutional layers, etc). But it is also true that generative models tend to underperform discriminative models in classification tasks. This will be particularly true in our implementation which uses flat priors.}  Ablation results are shown in Fig.~\ref{onehot ablation}. These are consistent with the previous experiments, with accuracy reducing (but still much better than chance value of 10\%) as the layers are ablated. In this case it would appear that the top 2 layers are adding nothing to the network's action selection ability. A key point to note is that the learning of the weights was not dependent on the provision of any information at the top of the network - all the learning comes about as a result of information presented at the bottom. Despite this, the network has distributed its ability through several layers, with the major part of successful inference relying on information towards the bottom of the network.

\begin{figure}
 \vspace{-0.2cm}
 \captionsetup{justification=raggedright}
\centering
\begin{subfigure}{.49\textwidth}
  \centering
  \includegraphics[width=.99\linewidth]{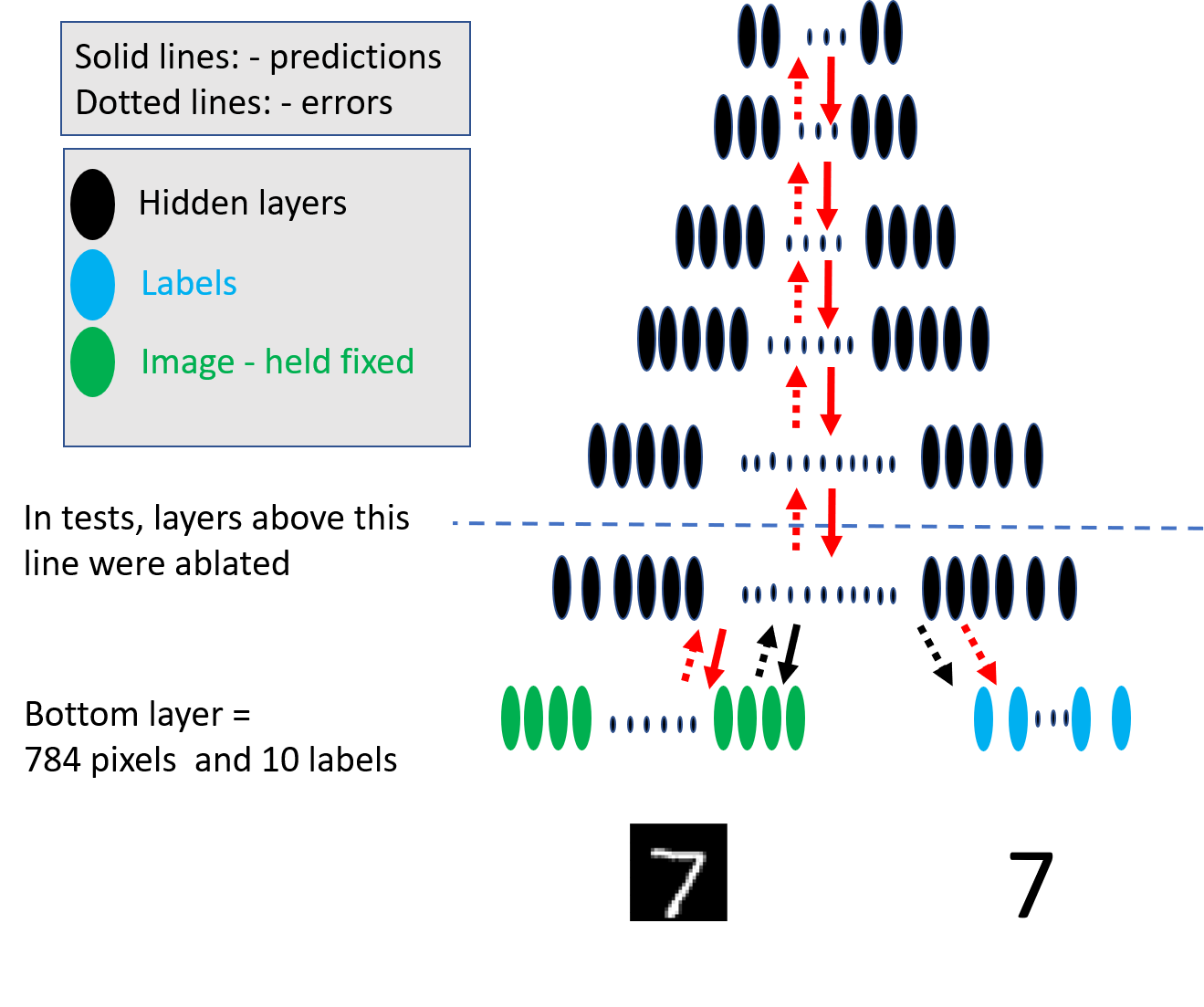}
  \caption{Schematic of network}
  \label{onehot network}
\end{subfigure}%
\begin{subfigure}{.49\textwidth}
  \centering
  \includegraphics[width=.99\linewidth]{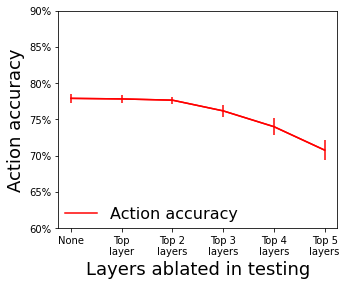}
  \caption{Effect of layer ablation}
  \label{onehot ablation}
\end{subfigure}%
\caption{A 7 layer PCN where 10 binary actions are associated with MNIST images. (a) Image and one-hot labels both at the bottom. For ease of reading, the nodes shown on each layer represent both value and error nodes. Red lines show flow of information with no ablation. Black line shows flow if 5 layers are ablated. (b) Ablation of top two layers has no effect on accuracy of action selection.  Ablation of the next 3 layers steadily reduces accuracy. Error bars are standard deviations across 10 differently seeded networks.}
\vspace{-0.6cm}
\label{onehot}
\end{figure}
\vspace{-0.4cm}

\section{Discussion}
\vspace{-0.2cm}
We have demonstrated that, when training a PCN with senses and actions at the bottom layer, it is not necessary to provide a high level "hidden state" in training in order to learn the correct actions for an incoming sensation. Furthermore, the network appears to distribute its learning throughout the layers, with higher layers called into use only as required. In our experiments, this meant that higher layers could be ignored if the lower layers alone contained sufficient information to select the correct action. In effect, the network has learned a sensorimotor shortcut to select the correct actions. On the other hand, if the higher layers contain information which improves action selection, then ablation of those layers reduces, but doesn't destroy, performance - ablation leads to graceful degradation. This flexibility is inherent in the nature of PCNs, unlike feed forward networks, which operate end to end. \\
\indent Importantly, this suggests that a PCN framework can help explain the development of fast reaction to a stimulus, even though the learning process involves all layers. 
For example, driving a car on an empty road might only require involvement of lower layers, whereas heavy traffic or icy conditions would require higher layers to deal with the more complex task. 
The fact that simple short-cuts can arise automatically during training and that the agent can dynamically select actions without involvement of higher layers could possibly also help explain why well-learned tasks can be carried out without conscious perception.\\  
\indent While we have provided an illustrative 'proof of principle' of this approach, much more can be done to investigate how this leads to a continuum of behaviour in active agents, which we list below in no particular order. Firstly, in our experiments inference took place with no influence from above and we have not considered the impact which exogenous priors would have. Secondly, we included no concept of a causal link between action and the subsequent sensory state. Action in real-life situations is a rolling process, with actions impacting subsequent decisions. 
Because our generative model did not consider time or state transitions, we cannot generalise to active inference in the sense of planning. One might argue that policy selection in active inference is a better metaphor for reflective behaviour, leading to a distinction between reflexive ‘homeostatic’ responses and more deliberative ‘allostatic’ plans. Having said this, it seems likely that the same conclusions will emerge. In other words, the same hierarchical generative model can explain reflective and reflexive behaviour at different hierarchical levels. 
Thirdly, the role of precisions has not been examined. Updating precisions should allow investigation of the role of attention. Finally, we have assumed the existence of a well trained network, and only touched on the performance of a partially trained network. It would be instructive to investigate how reliance on higher layers changes during the learning process.\\  
\indent These results support the view that a predictive coding network in the brain does not need to work from end to end, and can restrict itself to the number of lower layers required for the task at hand, possibly only in the sensorimotor system. There is the possibility of some tentative links here with more enactivist theories of the brain which posit that "representations" encode predicted action opportunities, rather than specify an abstract state of the world, but much further analysis is needed to investigate possible overlaps.

\section{Acknowledgements}
\vspace{-0.2cm}
PK would like to thank Alec Tschantz for sharing the "Predictive Coding in Python" codebase \url{https://github.com/alec-tschantz/pypc} on which the experimental code was based. Thanks also to three anonymous reviewers whose comments helped improve the clarity of this paper, particularly in relation to temporal aspects of predictive coding.  PK is funded by the Sussex Neuroscience 4-year PhD Programme. CLB is supported by BBRSC grant number BB/P022197/1.   

\bibliographystyle{splncs04}
\bibliography{thebibliography}

\appendix
\section{Network parameters}
\label{appendix:network_params}

 \textbf{Network size}: 4 layer

\noindent  \textbf{Number of nodes on each layer}: 10, 100, 300, 785 for MNIST-group and MNIST-digit1. 20, 100, 300, 785 for MNIST-barred. In the bottom layer, 784 nodes were fixed to the MNIST image, the 785\textsuperscript{th} node was an action node which updates in testing. In initial set of experiments, top layer was fixed to a one-hot representation of MNIST label in training.  In second set of experiments this was set to random value and allowed to update.

\noindent  \textbf{Non-linear function}: tanh

\noindent  \textbf{Bias used}: yes

\noindent  \textbf{Training set size}: full MNIST training set of 60,000 images, in batches of 640

\noindent  \textbf{Number of training epochs}: 10

\noindent  \textbf{Testing set size}: 1280 images selected randomly from MNIST test set

\noindent  \textbf{Learning parameters used in weight update of EM process}: Learning Rate=  1e-4, Adam

\noindent  \textbf{Learning parameters used in node update of EM process}: Learning Rate=  0.025, SGD

\noindent  \textbf{Number of SGD iterations in training}: 200

\noindent  \textbf{Number of SGD iterations in test mode}: 200 * epoch number.  The size is increased as epochs progress to allow for the decreasing size of the error between layers (as discussed in the text, this would normally be counteracted by increase in precision values).

\noindent  \textbf{Random initialisation}:  Except where fixed, all nodes were initialized with a random values selected from \(\mathcal{N}(0.5, 0.05)\)

\bigskip
\noindent In the experiment using a 7 layer network, the number of nodes on each layer were: 10, 25, 50, 100, 200, 300, 794. All other parameters the same as above

\end{document}